\documentclass[11pt,a4paper]{article}

\usepackage[T1]{fontenc}
\usepackage[utf8]{inputenc}
\usepackage[margin=1in]{geometry}
\usepackage{amsmath}
\usepackage{graphicx}
\usepackage{booktabs}
\usepackage{array}
\usepackage{multirow}
\usepackage[hidelinks]{hyperref}

\title{LLM-as-an-Investigator: Evidence-First Agentic Reasoning for Robust Interactive Problem Diagnosis}
\author{%
Fabrizio Marozzo\\
University of Calabria\\
Rende, Italy\\
\texttt{fabrizio.marozzo@unical.it}
\and
Pietro Li\`o\\
University of Cambridge\\
Cambridge, United Kingdom\\
\texttt{pietro.lio@cl.cam.ac.uk}
}
\date{}

\begin{document}
\maketitle

\begin{abstract}
Large language models (LLMs) are increasingly used for technical troubleshooting, where initial problem descriptions are often incomplete and several causes can explain the same symptoms. Effective diagnosis requires acquiring discriminative evidence and selecting the explanation best supported by it. However, additional reasoning or interaction does not necessarily prevent premature commitment to an incorrect hypothesis.

We introduce \textit{LLM-as-an-Investigator}, an evidence-first methodology for interactive problem diagnosis that reduces premature commitment while actively distinguishing competing causes. We implement it through the \textit{Solution Investigator Agent} (SIA), which orchestrates three complementary reasoning behaviours through a shared investigation state: a \textit{Diagnostic Reasoner} that conducts and reassesses the diagnostic interview, a \textit{Hypothesis Investigator} that maintains and updates competing hypotheses, and a \textit{Counterfactual Investigator} that challenges current explanations and guides discriminative follow-up questions. Final selection uses cited dialogue evidence and a separate verification step, while explicit hypothesis and evidence records make the diagnosis traceable and preserve unresolved alternatives.

We construct a structured corpus from 303 technical forum threads across mechanical, electrical, and hydraulic domains. A source-grounded pipeline separates the initial problem, pre-solution clarifications, and reported outcome, while a user simulator answers diagnostic questions without access to the reference solution. We compare SIA with direct, reasoning-based, interactive, and search-based baselines, including adaptations of ReAct and Tree of Thoughts. Evaluation distinguishes final diagnostic quality from hypothesis coverage and oracle potential, while ablations and paired stress tests assess component contributions and robustness to plausible but misleading user hypotheses.
\end{abstract}


\section{Introduction}
\label{sec:introduction}

Large language models (LLMs) have become a natural interface for technical problem solving. Users increasingly rely on conversational systems to diagnose software errors, debug devices, analyze failures, and obtain suggestions for resolving practical issues. Their ability to process informal natural language descriptions makes them useful for both expert and non-expert users \cite{brown2020language,ouyang2022training}. More recently, LLMs have become core components of agentic systems, combining reasoning with planning, memory, interaction, and control mechanisms \cite{wang2023surveyagents,luo2025llmagent}.

However, LLM-based assistants remain vulnerable when the initial problem description is incomplete or ambiguous. Such queries often require explicit clarification, yet models may answer directly when asking a question would be more appropriate \cite{kuhn2022clam,kim2023tree,ijns-2026}. In technical troubleshooting, several causes can explain the same symptoms, and a plausible initial diagnosis may become an anchor for subsequent reasoning. Unverified user suggestions can further influence this process if they are accepted without testing alternatives \cite{perez2022discovering,sharma2023towards,ranaldi2023when,ozturk2026generative}.

Premature diagnoses can lead to unnecessary repairs, wasted time, and loss of trust. Related reliability concerns arise when LLMs produce plausible but unsupported content \cite{li2023halueval} or reinforce incorrect user assumptions \cite{bo2025invisible}. For example, a pump that starts intermittently but operates normally once running may initially be attributed to a defective pressure switch. Additional observations about its startup behavior may instead favor a fault in the motor's starting circuit, such as a degraded capacitor. The challenge is both to acquire information that distinguishes competing explanations and to select the diagnosis best supported by it.

This paper proposes \textit{LLM-as-an-Investigator}, an evidence-first agentic AI methodology implemented through the \textit{Solution Investigator Agent} (SIA). SIA organizes the investigation around three specialized reasoning behaviours: diagnostic reasoning, structured hypothesis investigation, and counterfactual investigation. Together, they support iterative diagnosis reassessment, explicit maintenance of competing causes, and targeted acquisition of evidence that distinguishes unresolved alternatives. Final selection retains the Diagnostic Reasoner's diagnosis unless cited dialogue evidence and a separate verification step support a revision.

To evaluate the approach, we organize case preparation and diagnostic evaluation as a multi-agent framework using a structured case corpus derived from 303 technical forum threads across mechanical, electrical, and hydraulic domains. The \textit{Problem-Solution Extractor Agent} identifies the initial problem, pre-solution clarifications, and reported solution. The \textit{Solution Investigator Agent} conducts the diagnosis by orchestrating the three specialized reasoning behaviours described above. During evaluation, the \textit{Grounded User Simulator Agent} simulates the user by answering diagnostic questions using the permitted context, without access to the reference solution. The extractor prepares the cases offline, while the simulator replaces the human user only during evaluation. This framework supports comparisons with direct prompting, non-interactive reasoning, interactive reasoning, a fixed-question baseline, and diagnostic adaptations of ReAct and Tree of Thoughts, alongside exploratory evaluations with different LLM backbones.

Evaluation distinguishes final diagnostic quality from hypothesis coverage and oracle potential: generating a suitable explanation does not imply selecting it. We analyze the recorded candidates and final diagnosis together with interaction cost. Component ablations assess architectural contributions, while paired stress tests examine sensitivity to plausible but misleading user hypotheses.

The main contributions are as follows:
\begin{itemize}
\item We propose SIA, an interactive diagnostic architecture coordinating three specialized reasoning behaviours through a shared investigation state: diagnostic reasoning, structured hypothesis investigation, and counterfactual investigation, with evidence-grounded final selection and explicit records explaining the selected diagnosis and preserving unresolved alternatives.
\item We construct a structured case corpus and a source-grounded three-agent evaluation pipeline across electrical, hydraulic, and mechanical domains, with explicit case provenance and reproducible execution protocols.\footnote{\url{https://github.com/SCAlabUnical/llm-as-an-investigator}}
\item We compare SIA with direct, interactive, and search-based baselines, distinguishing diagnostic quality from hypothesis coverage, and examine component contributions, backbone sensitivity, and robustness to misleading user hypotheses.
\end{itemize}

The remainder of the paper is organized as follows. Section~\ref{sec:related} reviews related work. Section~\ref{sec:methodology} describes SIA. Section~\ref{sec:experiments} presents the dataset, evaluation protocol, and results. Section~\ref{sec:conclusion} concludes the paper.

\section{Related Work}
\label{sec:related}

\subsection{Prompting and Reasoning in Large Language Models}

Large language models (LLMs) can perform diverse tasks through natural-language instructions without modifying their parameters. GPT-3 demonstrated in-context learning through zero-shot, one-shot, and few-shot prompting \cite{brown2020language}, motivating research on prompt engineering for more reliable and task-specific behavior \cite{sahoo2024systematic}.

Chain-of-Thought (CoT) prompting elicits intermediate reasoning steps before a final answer \cite{wei2022chain}, while Zero-Shot CoT achieves similar behavior without demonstrations \cite{kojima2022large}. Self-consistency aggregates answers from multiple sampled reasoning trajectories \cite{wang2022self}. Other methods decompose complex tasks into simpler subproblems, as in least-to-most prompting \cite{zhou2022least}, or automatically construct reasoning demonstrations, as in Auto-CoT \cite{zhang2022automatic}.

Beyond linear reasoning, Tree of Thoughts explores and evaluates alternative reasoning branches \cite{yao2023tree}, while ReAct interleaves reasoning with actions and external observations \cite{yao2023react}. These approaches provide mechanisms for deliberation, exploration, and interaction. Technical diagnosis additionally requires deciding which missing evidence to acquire, maintaining alternative causes as the dialogue evolves, and determining whether the collected observations support a final diagnosis.

\subsection{Interactive Diagnosis and Information Seeking}

Clarification and adaptive information gathering are established problems in interactive language-model systems. Zhang and Choi distinguish three decisions: when clarification is needed, what question to ask, and how to use the response \cite{zhang2025clarify}. In clinical reasoning, MediQ combines a simulated patient with an expert system that can defer diagnosis and request missing information \cite{li2024mediq}. Its results show that directly prompting LLMs to ask questions can degrade performance, motivating explicit strategies for deciding when to continue the interaction.

Sequential Diagnosis with Language Models introduces a benchmark of stepwise clinical encounters and the MAI Diagnostic Orchestrator (MAI-DxO) \cite{nori2025sequential}. The orchestrator coordinates roles for generating differential diagnoses, selecting informative tests, challenging the leading explanation, and controlling diagnostic costs. Its \textit{Dr. Challenger} explicitly addresses anchoring bias by seeking contradictory evidence and proposing falsifying tests. Thus, maintaining alternatives and challenging an initially plausible diagnosis are already important components of prior diagnostic systems.

BED-LLM formulates adaptive information gathering through sequential Bayesian experimental design, selecting questions that maximize expected information gain under an explicit probabilistic model \cite{choudhury2025bedllm}. MedClarify applies information-seeking diagnosis to medicine, maintaining candidate diagnoses and selecting follow-up questions according to their expected information gain \cite{wong2026medclarify}. These methods provide principled uncertainty-based question selection; SIA instead uses diagnostic disagreements and counterfactual challenges to guide follow-up questioning.

ALFA improves question generation through preference-based alignment over fine-grained attributes, including clarity, answerability, diagnostic relevance, and avoidance of differential-diagnosis bias \cite{li2025alfa}. This complements approaches that organize the diagnostic process through an external controller. More recently, EviMed evaluates evidence elicitation in interactive diagnosis, showing that strong reasoning does not necessarily imply effective information collection \cite{long2026evidence}. The same work introduces REFINE, which uses diagnostic verification to identify evidence gaps and guide subsequent questions.

These studies establish that effective diagnosis requires coordinated reasoning, evidence acquisition, and verification. Our work builds on this foundation and examines a particular organization of these processes in technical troubleshooting. SIA combines a reasoning-led initial interview with a separate structured investigation of the collected dialogue, followed by counterfactual questioning and evidence-grounded final selection.

\subsection{Sycophancy and User-Driven Bias in LLM Interaction}

Instruction tuning and reinforcement learning from human feedback improve instruction following \cite{ouyang2022training}, but assistants may also align with unsupported user beliefs. Perez et al. identified sycophancy-related behaviors in language models \cite{perez2022discovering}, while Sharma et al. showed that human preferences can favor responses agreeing with users over truthful alternatives \cite{sharma2023towards}. Ranaldi and Pucci examined susceptibility to human-influenced prompts \cite{ranaldi2023when}, and SycEval evaluated responses to user rebuttals across commercial models \cite{fanous2025syceval}.

Such agreement can affect problem solving: \textit{Invisible Saboteurs} studies how sycophantic assistants can mislead novice users \cite{bo2025invisible}. Reasoning can mitigate biased final decisions, but can also produce justifications that mask sycophantic behavior \cite{feng2026good}. In technical diagnosis, an unverified user explanation is therefore a candidate cause to investigate rather than an observation supporting that cause. We assess this issue through controlled tests with misleading user hypotheses, as an additional robustness dimension of interactive diagnosis.

SIA does not claim to introduce hypothesis-based diagnosis or counterfactual verification. Its design combines hierarchical candidate causes, cautious plausibility updates, evidence-supported hypothesis expansion, and verification of proposed diagnostic revisions. Our evaluation distinguishes the coverage of the recorded hypothesis set from the quality of the final selected diagnosis, and compares SIA with direct, interactive, and search-based baselines under explicitly reported interaction limits.

\section{Methodology}
\label{sec:methodology}

\begin{figure}[t]
\centering
\includegraphics[width=0.75\textwidth]{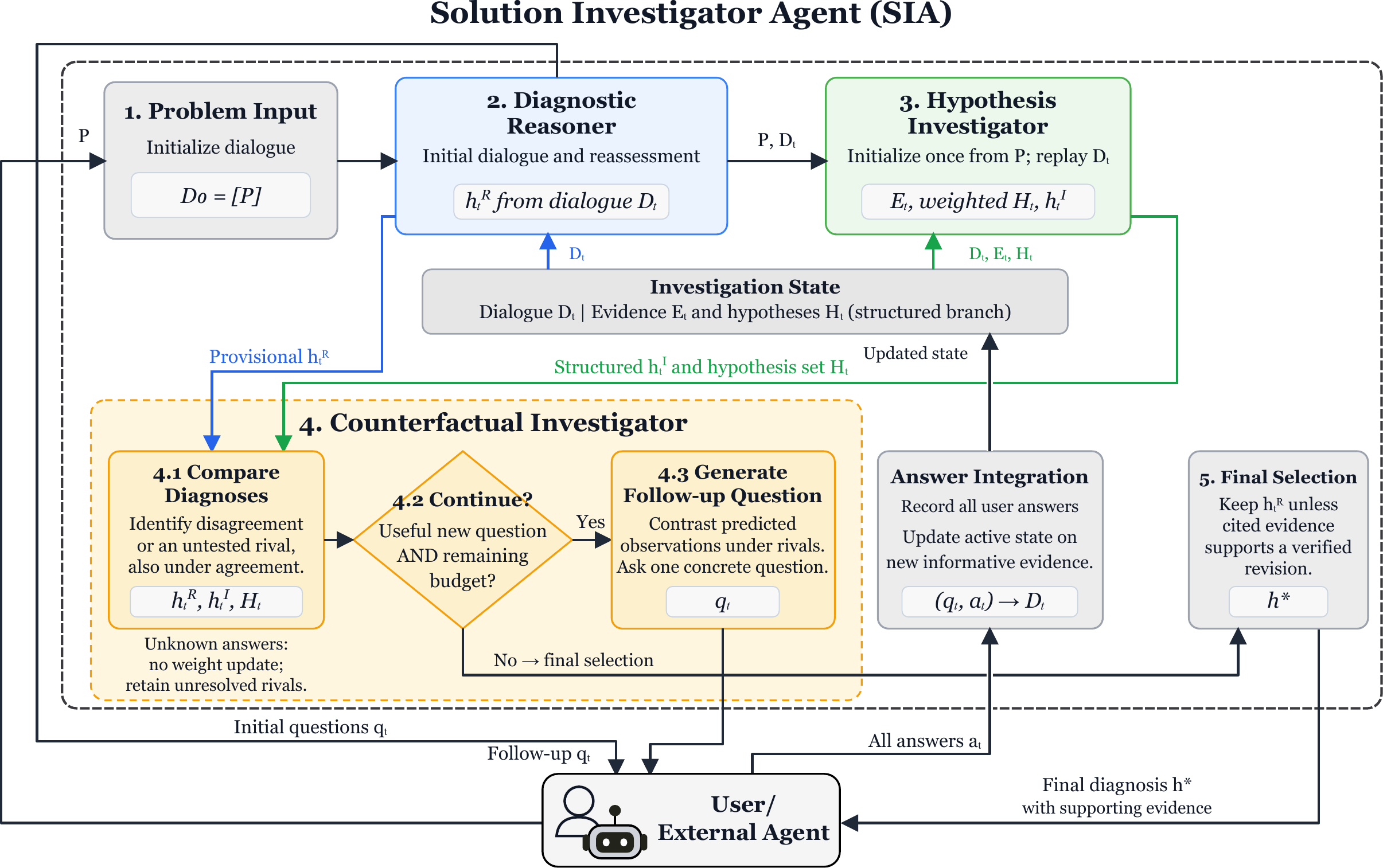}
\caption{SIA architecture and investigation flow. SIA orchestrates three specialized reasoning behaviours---Diagnostic Reasoner, Hypothesis Investigator, and Counterfactual Investigator---through a shared investigation state. Numbered blocks denote the investigation sequence; behaviours are re-invoked as new evidence is collected until final selection.}
\label{fig:methodology}
\end{figure}

This section describes how the \textit{Solution Investigator Agent} (SIA) investigates a technical problem by orchestrating specialized reasoning behaviours for clarification, hypothesis analysis, counterfactual investigation, and evidence-supported diagnostic selection. Figure~\ref{fig:methodology} summarizes their organization and investigation flow.

We use a running example of an apartment water pump that intermittently fails to start, with plausible causes spanning hydraulic, control, and motor-starting faults. The reported cause is a degraded starting capacitor, which is not available to SIA during investigation; the examples below illustrate individual reasoning steps rather than a complete execution trace.

\subsection{Architecture and Investigation State}

As shown in Fig.~\ref{fig:methodology}, SIA orchestrates three complementary reasoning behaviours through a shared investigation state. The \textit{Diagnostic Reasoner} conducts the initial interview and reassesses its diagnosis from the recorded dialogue. The \textit{Hypothesis Investigator} maintains explicit candidate causes, examines their compatibility with observations, and updates the structured hypothesis state. The \textit{Counterfactual Investigator} challenges current explanations and guides discriminative follow-up questioning. These behaviours are implemented through separate LLM calls coordinated by the controller rather than as independently deployed agents.

We denote the initial problem by $P$, the chronological dialogue by $D_t$, the evidence record by $E_t$, and the structured hypothesis set by $H_t$. The current diagnoses proposed by the \textit{Diagnostic Reasoner} and the \textit{Hypothesis Investigator} are denoted by $h_t^R$ and $h_t^I$, respectively. A question $q_t$ and its answer $a_t$ are recorded in the dialogue, which initially contains only $P$. The final diagnosis is denoted by $h^*$. The structured hypothesis state is initialized after the initial interview; \textit{Answer Integration} records answers and makes them available to the active reasoning phase.

Each structured hypothesis has an identifier, descriptive label, fault domain, subsystem, component, failure mechanism, and plausibility weight. This hierarchy allows SIA to move from broad fault families to specific causes while retaining alternative explanations in its trace.

The initial structured hypothesis set is generated solely from the problem description $P$. The answers collected during the initial interview are then incorporated in chronological order to update this set. Its size depends on the distinct plausible mechanisms identified by the model, rather than a fixed number of hypotheses. The initialization considers broad fault families relevant to the problem domain before proposing concrete causes. An \textit{other cause} candidate represents explanations missing from the list, but cannot serve as the final concrete diagnosis. Initial weights are validated and normalized by the controller. They are heuristic plausibility values, not calibrated probabilities, and do not alone determine the final choice.

SIA separates semantic reasoning from state management and execution control \cite{wang2023surveyagents,zhao2023indepthagents}. LLM calls propose questions, analyze observations, and compare explanations. The controller records the interaction, validates structured responses and evidence citations, maintains candidate weights, and enforces the question budget.

\subsection{Interactive Investigation}

\paragraph{Initial reasoning-led interview.}
The \textit{Diagnostic Reasoner} receives the problem description $P$ and conducts an initial interview, asking one diagnostic question at a time. It considers plausible causes and seeks observations that could distinguish competing explanations or clarify missing details. Each answer is added to the dialogue and informs the next question. The interview stops when its allocated question capacity is exhausted or when the reasoner judges that no further question would be useful. It then reviews the collected dialogue, proposes one diagnostic cause, and records the alternatives explicitly considered.

The reasoner remains active during subsequent hypothesis-driven investigation. Whenever a follow-up answer provides new informative evidence, it reassesses its diagnosis using the updated dialogue; its current explanation may therefore be retained, refined, or replaced. These assessments use the problem description and the actual question--answer history, without receiving the structured investigator's hypotheses or plausibility weights. The two diagnosis-producing behaviours can consequently propose different diagnoses, allowing SIA to investigate their disagreement through counterfactual questioning.

A shared budget $B$ limits all questions, with unused initial-interview capacity available for subsequent hypothesis-driven questioning; its experimental setting is reported in Section~\ref{sec:experiments}.

\paragraph{Structured analysis and counterfactual critique.}
After initialization from $P$, the \textit{Hypothesis Investigator} processes the initial question--answer pairs chronologically. Before analyzing each answer, the \textit{Counterfactual Investigator} considers the current leading hypothesis and proposes an observable finding that would weaken it, together with a corresponding question. This encourages consideration of contradictory evidence. A proposed falsifier remains hypothetical: it is never recorded as an observation unless an actual answer provides that information.

Answer Integration records all responses, including unavailable information. Weight updates require new informative evidence; benchmark provenance annotations identify answer support without exposing the reference diagnosis. Unknown or non-informative answers leave the weights unchanged.

For each new informative answer, the model estimates its compatibility with each current hypothesis $i$, whose plausibility weight is denoted by $w_i^{(t)}$. The controller converts these estimates into positive update factors $\phi_i^{(t)}$, limiting their relative contrast and tempering their effect to avoid abrupt changes driven by a single answer. It then updates and normalizes the plausibility weights:
\begin{equation}
w_i^{(t)} = \frac{w_i^{(t-1)}\phi_i^{(t)}}{\sum_j w_j^{(t-1)}\phi_j^{(t)}}.
\label{eq:sia_weight_update}
\end{equation}
The update applies to the hypotheses present before any expansion at that step. These weights express heuristic plausibility rather than a calibrated Bayesian posterior. An incomplete compatibility response triggers a repair request; if it remains incomplete, the update is skipped and the failure is logged.

In the pump example, a persistent obstruction may initially be considered. The \textit{Counterfactual Investigator} identifies normal water delivery after startup as an observation that would weaken this explanation. If the user confirms that the pump operates normally once started, this observation can reduce support for a persistent obstruction relative to startup-related causes, without establishing a specific component failure. If the user cannot answer, the plausibility weights remain unchanged.

\paragraph{Evidence-supported hypothesis expansion.}
The candidate set remains open to revision. After an informative answer, the \textit{Hypothesis Investigator} checks whether the new observation supports a distinct concrete cause missing from the current set. A proposed addition must identify a component and cite the supporting answer through its turn index and a quotation. The controller validates the citation and rejects duplicate component labels, unspecified placeholders, and descriptions that combine alternative causes.

Each expansion step admits at most one new hypothesis. The controller assigns it an initial plausibility weight and rescales the existing weights while preserving their relative proportions; the initialization setting is reported in the experimental protocol. The proposed cause need not be explicitly named by the user: an observed operating pattern may support a component-level explanation. Its inclusion records a possibility supported by that observation, rather than establishing it as the correct diagnosis.

\paragraph{Disagreement and rival-driven questioning.}
After processing the initial dialogue, the \textit{Hypothesis Investigator} makes a fresh diagnosis from all actual answers and its candidate set. It reviews explanations even when their heuristic weights are low. SIA compares this diagnosis with the \textit{Diagnostic Reasoner}'s current diagnosis at the component level. If they disagree, the \textit{Counterfactual Investigator} seeks an observation that distinguishes their explanations. If they agree, the controller looks for a concrete rival that has not yet received an informative challenge against the agreed diagnosis. Agreement alone therefore does not end questioning when such a rival remains.

The \textit{Counterfactual Investigator} considers the dialogue and competing causes, contrasts observations expected under rival explanations, and asks one concrete question about something the user may already know from ordinary operation. In the evaluated setting, it does not request a new physical test, ask for the reference solution, or assume an unperformed test has occurred. Question selection is based on model reasoning about discriminative observations; it does not optimize a numerical expected-information-gain objective.

An informative follow-up answer updates the structured state and can support hypothesis expansion. The \textit{Diagnostic Reasoner} also reassesses its diagnosis from $P$ and the actual dialogue, without receiving the structured candidate list or weights. The \textit{Hypothesis Investigator} reviews the updated state before the next comparison. An unavailable answer neither changes the weights nor marks the challenged rival as resolved; further questions remain possible within the remaining budget. Thus, the three reasoning behaviours can contribute repeatedly as the investigation evolves.

In the pump example, the reasoner may favor a pressure-switch fault while the investigator proposes a degraded starting capacitor. SIA can ask whether, during a failed start, the motor is silent or hums without rotating. A reported hum can weaken the explanation that the motor receives no power, but does not uniquely identify a capacitor fault. Even if both diagnostic behaviours favor the capacitor, a mechanical starting impediment may remain a rival, motivating a further question about any previously observed unusual noise or irregular operation after startup. Neither observation establishes the cause on its own; unavailable information leaves the competing explanations unresolved.

\subsection{Stopping and Final Selection}

Investigation stops when the question budget is exhausted, the \textit{Counterfactual Investigator} reports that no useful follow-up remains, no valid new question can be generated, or the \textit{Diagnostic Reasoner} and \textit{Hypothesis Investigator} agree and the controller finds no remaining concrete rival to challenge. Empty questions and exact textual repetitions are rejected. Stopping does not depend on a fixed confidence threshold over the plausibility weights, and budget exhaustion does not imply diagnostic certainty.

Final selection uses the \textit{Diagnostic Reasoner}'s diagnosis as the default. The \textit{Hypothesis Investigator} may replace it only after consistency checks and verified dialogue evidence. Its proposed cause must be internally consistent and supported by a cited informative answer or a cited observation in $P$. The final review may introduce a missing component only when supported by cited evidence; such an addition remains a candidate rather than directly determining the final diagnosis. Otherwise, SIA retains the reasoner's diagnosis.

When the diagnoses differ, an adjudicator compares anonymized alternatives A and B using $P$ and the dialogue. A revision requires a cited observation that materially favors the investigator's component and passes separate evidence verification. SIA therefore returns one diagnosis $h^*$ while preserving unresolved alternatives in the trace.

\subsection{Robustness and Traceability}

SIA explicitly separates user conjectures from observed problem information before investigation. For example, ``the pump hums but does not start'' reports an observation, whereas ``I think the non-return valve is faulty'' proposes a cause. Such conjectures may be retained as hypotheses for subsequent comparison, but are not treated as established facts by the reasoner or the simulated user. A suggested cause can influence the final diagnosis only when supported by a cited observation from $P$ or an informative answer and validated through a separate verification step. This mechanism allows user suggestions to be considered without treating them as evidence by default.

SIA returns one selected diagnosis, the recorded candidate causes, and the investigation trace. Logs include questions and answers, evidence annotations, weight updates, hypothesis additions, diagnostic proposals, and final selection checks. These records allow final diagnostic quality to be evaluated separately from hypothesis coverage: considering a correct explanation and selecting it are distinct outcomes.

All reasoning roles use API-based LLM calls without local training or fine-tuning. Structured outputs and diagnostic fields are validated by the controller; failures remaining after the applicable repair steps are recorded. Prompts, configuration settings, and execution traces support reproducibility. Model configurations, interaction budgets, and evaluation procedures are reported in Section~\ref{sec:experiments}.

\section{Experimental Evaluation}
\label{sec:experiments}

\begin{figure}[htb!]
\centering
\includegraphics[width=0.85\textwidth]{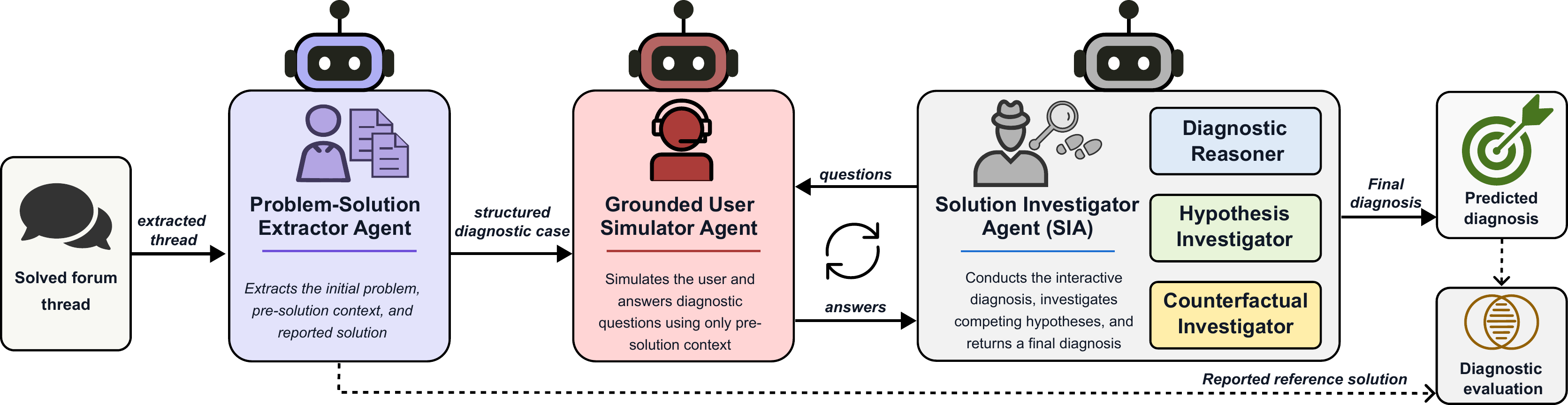}
\caption{Three-agent evaluation pipeline. The Problem-Solution Extractor Agent constructs diagnostic cases, the Grounded User Simulator Agent answers questions without access to the reference solution, and SIA orchestrates its diagnostic, hypothesis-investigation, and counterfactual behaviours to produce a final diagnosis evaluated separately against the reported outcome.}
\label{fig:evaluation_pipeline}
\end{figure}

We compare seven techniques in terms of final diagnostic quality, hypothesis coverage, and interaction cost. Repeated executions assess variability, while additional studies examine component contributions, misleading user suggestions, and alternative LLM backbones. Figure~\ref{fig:methodology} details the specialized reasoning behaviours orchestrated within SIA, whereas Figure~\ref{fig:evaluation_pipeline} places SIA within the broader three-agent evaluation pipeline together with the Problem-Solution Extractor Agent and the Grounded User Simulator Agent.

\subsection{Experimental Setup}
\label{sec:experimental_setup}

The evaluation follows the three-agent pipeline in Figure~\ref{fig:evaluation_pipeline}. The source corpus comprises 303 technical forum threads and 8,930 posts: 96 electrical, 105 hydraulic, and 102 mechanical threads, averaging 29.5 posts per thread. In these discussions, users describe a technical problem and exchange messages with other participants, including experienced users, to clarify symptoms, explore possible causes, and identify a solution. We selected resolved threads with a reported outcome. The \textit{Problem-Solution Extractor Agent} converts each thread into a structured case containing the initial problem, pre-solution clarifications, and reported solution. Source quotations and post ordering separate available observations from outcome-revealing information. Reported solutions serve as diagnostic references rather than independently verified physical ground truth. Source threads, structured cases, and code are available in the project repository.\footnote{\url{https://github.com/SCAlabUnical/llm-as-an-investigator}}

The comparative cohort contains 100 source-validated cases: 36 electrical, 30 hydraulic, and 34 mechanical. These cases were selected because they provided sufficiently rich diagnostic information and a clearly identifiable reported solution, making them suitable for reliable comparative evaluation. All techniques receive the same initial problem. Interactive methods question the \textit{Grounded User Simulator Agent}, which receives only permitted pre-solution information, without access to the reference solution. It answers from explicit observations or supported inferences, verifies inferred answers through a separate model call, and reports unsupported information as unavailable. Final diagnoses are evaluated separately against the references.

The main evaluation uses Gemini 3.8 Flash for the investigation backbone, the \textit{Grounded User Simulator Agent}, and the automatic evaluator. For backbone sensitivity, we repeat the SIA--THK-int comparison by replacing only the investigation backbone with GPT-5.6 Terra or Claude Sonnet 5, while retaining the same prompts, investigation logic, question budget, structured-output constraints, simulator, evaluator, and evaluation protocol. Each method is executed three times per case. Runs are first averaged within each case, and we report mean scores and 95\% confidence intervals via percentile bootstrap over cases using 10,000 resamples.

Adaptive interactive methods operate under a maximum question budget of $B=20$ and may terminate earlier; Ask-5 is capped at five questions by design. Plausibility updates use $\epsilon=0.001$ to prevent zero factors, $r=4$ to bound compatibility contrast, and $\eta=0.5$ to temper updates; together, the latter two settings limit the ratio between update factors to two for a single observation. Answer-supported additions receive weight $0.15$, with existing weights proportionally rescaled, while final-review additions receive an unnormalized weight of $0.05$ before normalization. These fixed engineering settings moderate belief updates and candidate insertion without treating the resulting weights as calibrated probabilities or imposing a fixed number of hypotheses.

Each technique must return one final cause. Invalid or non-unique diagnoses receive zero and remain in the denominator. Technical and schema failures are recovered with unchanged settings and retained attempts; valid or ambiguous diagnoses are not regenerated to improve scores. Logs preserve hypotheses, dialogue, evidence annotations, decisions, model settings, durations, and API usage.

\subsection{Compared Techniques and Metrics}
\label{sec:baselines_metrics}

We compare the proposed \textbf{SIA} method against direct, reasoning-based, interactive, and search-based diagnostic baselines:

\begin{itemize}
    \item \textbf{BAS}: produces a direct diagnostic response from the initial problem without requesting additional information; alternative causes explicitly proposed in the response are retained as recorded candidates.

    \item \textbf{THK}: analyzes competing explanations before selecting a diagnosis, without interacting with the user.

    \item \textbf{THK-interactive (THK-int)}: combines explicit reasoning with adaptive clarification questions and revises its diagnosis from the collected answers.

    \item \textbf{Ask-5}: asks up to five sequential clarification questions, with each question conditioned on previous answers, before selecting a diagnosis.

    \item \textbf{ReAct}: our diagnostic adaptation of ReAct~\cite{yao2023react}, alternating explicit analysis, clarification actions, and simulator observations until a final diagnosis is produced.

    \item \textbf{Tree of Thoughts (ToT)}: our diagnostic adaptation of Tree of Thoughts~\cite{yao2023tree}, which generates, evaluates, and prunes alternative diagnostic reasoning paths; clarification answers are shared observations used across surviving paths.
\end{itemize}

ReAct and ToT are adapted to the diagnostic setting, with all interactive methods evaluated under the same question budget.

A common automatic evaluator applies a hierarchical semantic rubric, followed by a causal-failure check. Scores range from 0 to 100, with 100, 75, 50, 25, and 0 serving as reference points rather than discrete categories. A score of 100 requires the correct component or fault location together with the correct causal failure; around 75 denotes the same subsystem, or a matching component with a different or unconfirmed failure; around 50 identifies the correct broad fault domain; around 25 denotes a partially related but insufficiently localized diagnosis; and 0 denotes an unrelated or invalid diagnosis. Intermediate scores reflect partial semantic and causal agreement within these levels. The evaluator receives neither the technique name nor the candidate position. These values are graded diagnostic scores, not percentages of correct diagnoses.

The \textit{final diagnosis score} evaluates the selected cause. The \textit{potential score} is the highest reference-based score among the recorded candidates, representing an oracle selector applied retrospectively. \textit{Exact hypothesis coverage} counts outcomes containing a candidate scoring 100; \textit{exact final diagnosis} counts selected causes scoring 100. \textit{Initial potential} applies the same oracle score to the recorded initial set, where available. Initial sets differ operationally between methods; a dash indicates that no separate initial set is recorded. Candidate positions, hypothesis counts, and question counts are retained. Positions refer to the recorded list, not a calibrated probabilistic ranking.


\subsection{Comparative Evaluation}
\label{sec:comparative_results}

Table~\ref{tab:gemini_comparison} reports case-averaged diagnostic scores and interaction statistics. SIA achieves the highest final mean, 79.2/100, compared with 72.8 for THK-int and 72.3 for Ask-5. Its mean advantage is 11.9 points over BAS, 11.7 over THK, 6.4 over THK-int, and 6.9 over Ask-5. ReAct and ToT obtain 69.0 and 61.5, respectively. These descriptive differences favor SIA on the evaluated cohort.

\begin{table}[thb!]
\centering
\small
\setlength{\tabcolsep}{4pt}
\caption{Comparative evaluation on 100 distinct cases. Scores are out of 100; 95\% confidence intervals bootstrap cases after averaging runs within each case. $H$ and $Q$ denote mean recorded hypotheses and questions.}
\label{tab:gemini_comparison}
\begin{tabular}{lrrrrr}
\toprule
Technique  & Initial pot. & Potential & Final [95\% CI] & $H$ & $Q$ \\
\midrule
BAS  & -- & 81.2 & 67.3 [59.3, 75.0] & 4.3 & 0.0 \\
THK  & -- & 81.0 & 67.5 [58.5, 76.0] & 4.5 & 0.0 \\
THK-int  & -- & 84.7 & 72.8 [66.8, 79.8] & 4.5 & 3.7 \\
Ask-5 & -- & 83.3 & 72.3 [64.7, 79.5] & 4.5 & 5.0 \\
ReAct  & 77.5 & 86.5 & 69.0 [60.5, 77.0] & 12.7 & 2.8 \\
ToT  & 77.0 & 84.0 & 61.5 [52.5, 70.0] & 21.0 & 2.5 \\
\textbf{SIA}  & \textbf{79.3} & \textbf{89.2} & \textbf{79.2} [72.7, 83.7] & 9.0 & 5.2 \\
\bottomrule
\end{tabular}
\end{table}

The table separates candidate generation from final selection. SIA has the highest recorded potential, 89.2/100, compared with an initial potential score of 79.3. Its final score remains 10.0 points below its potential, showing that considering a suitable explanation does not ensure its selection. ToT exhibits a larger gap, with a potential of 84.0 and a final score of 61.5. Broader exploration therefore does not necessarily produce better diagnostic selection.

The potential score represents an upper bound for an ideal selector restricted to the recorded candidate pool and given access to the reference solution. It is not necessarily attainable from the available dialogue: if the discriminating observation is missing, the correct cause may be present among the candidates without being identifiable with certainty. SIA's explicit hypothesis and evidence records make the selection process inspectable, documenting why a cause is chosen and which alternatives remain unresolved. The potential--final gap alone cannot distinguish missing evidence from selection errors.

Coverage also depends on the number and visibility of recorded candidates. ReAct and ToT expose intermediate proposals, including pruned paths, whereas the simpler baselines predominantly record their final alternatives. This logging asymmetry limits direct interpretation of coverage differences. SIA records 9.0 hypotheses and asks 5.2 questions on average, compared with 21.0 hypotheses for ToT and 3.7 questions for THK-int. Its higher diagnostic score is accompanied by more questions than the other interactive methods, making interaction cost relevant alongside diagnostic quality.

Separate backbone evaluations show the same relative advantage despite lower absolute performance than with Gemini. With GPT-5.6 Terra, SIA reaches a final diagnostic score of 75.4 with a potential of 85.1, compared with 69.2 and 80.6 for THK-int. With Claude Sonnet 5, SIA obtains 76.1 final and 86.0 potential, compared with 68.4 and 81.2 for THK-int. Thus, although both alternative backbones exhibit lower absolute final and candidate-pool potential scores than Gemini, SIA preserves margins of 6.2 and 7.7 points in final diagnostic quality, respectively.

\subsection{Ablation Study}
\label{sec:ablation}

We evaluate the contribution of the main SIA components on the same 100 source-validated cases and under the same experimental setting used in the comparative evaluation. The analysis removes structured investigation, counterfactual critique, evidence-based weight updates, hypothesis expansion, follow-up questioning, and final adjudication. Table~\ref{tab:ablation} reports the resulting final diagnostic and potential scores.

\begin{table}[thb!]
\centering
\small
\setlength{\tabcolsep}{3pt}
\caption{Component ablation of SIA. Final denotes the selected diagnostic score, while Potential denotes the oracle score over the recorded candidate pool.}
\label{tab:ablation}
\begin{tabular}{lrr}
\toprule
Configuration & Final & Potential \\
\midrule
Full SIA & 79.2 & 89.2 \\
Without structured investigation & 72.8 & 84.7 \\
Without counterfactual critique & 74.0 & 85.5 \\
Without weight updates & 78.5 & 88.5 \\
Without hypothesis expansion & 76.5 & 79.3 \\
Without follow-up questioning & 75.0 & 85.0 \\
Without final adjudication & 76.0 & 89.2 \\
\bottomrule
\end{tabular}
\end{table}

Removing structured investigation produces the largest decrease in final diagnostic quality, from 79.2 to 72.8, while the potential score decreases from 89.2 to 84.7. Counterfactual critique also has a substantial impact: disabling it reduces the final score to 74.0 and the potential score to 85.5. These results support the role of counterfactual challenges in testing leading explanations and directing the investigation toward discriminative evidence.

Disabling follow-up questioning reduces the final score to 75.0 and the potential score to 85.0, indicating that additional interaction contributes to both candidate identification and final selection. Hypothesis expansion has the strongest effect on candidate-pool potential: without it, the potential score drops from 89.2 to 79.3, while the final score decreases to 76.5. This suggests that evidence-supported expansion is particularly important for recovering plausible causes that are not sufficiently represented in the initial hypothesis set.

Evidence-based weight updates have a smaller individual effect, with final and potential scores decreasing to 78.5 and 88.5, respectively. Their role is therefore mainly to refine the relative plausibility and prioritization of already identified hypotheses.

Finally, removing final adjudication leaves the potential score unchanged at 89.2 but reduces the final score to 76.0. This indicates that final adjudication primarily improves selection among already available candidates rather than candidate generation. Overall, the ablation results highlight complementary roles across the architecture: structured investigation and counterfactual critique strongly affect diagnostic quality, hypothesis expansion broadens candidate coverage, follow-up questioning supports evidence acquisition, weight updates refine prioritization, and final adjudication improves the final selection.

\subsection{Robustness to Misleading User Suggestions}
\label{sec:robustness}

We evaluate whether SIA remains robust when the user introduces a plausible but incorrect explanation of the problem~\cite{kim2025challenging,sharma2023towards,fanous2025syceval}. We select 10 source-validated cases and evaluate each under two paired conditions, resulting in 20 diagnostic inputs. In the \textit{control} condition, SIA receives the original problem description; in the \textit{challenged} condition, the same description is augmented with a plausible but incorrect first-person conjecture about the cause. All remaining context, simulator information, references, and experimental settings are unchanged.

The robustness subset is not a random sample of the full cohort, as cases must support a plausible alternative explanation and sufficiently explicit diagnostic evidence. This may make them somewhat easier than the overall corpus, so we focus on paired control--challenged differences rather than aggregate main-study scores. Because the paired conditions differ only in the added conjecture, the design directly measures sensitivity to user-supplied causal framing while controlling for the underlying diagnostic case.

SIA separates user conjectures from observed evidence. A proposed cause may remain among the considered hypotheses, but it is not treated as supporting evidence unless subsequent observations and verification provide independent support.

\begin{table}[t]
\centering
\small
\setlength{\tabcolsep}{4pt}
\caption{Robustness to misleading user suggestions on 10 paired cases. The challenged condition adds a plausible but incorrect user conjecture.}
\label{tab:sycophancy}
\begin{tabular}{lcc}
\toprule
Measure & Control & Challenged \\
\midrule
Final diagnostic score & 80.0 & 76.0 \\
Potential score & 90.0 & 85.0 \\
Exact final diagnosis & 6/10 & 5/10 \\
Exact hypothesis coverage & 8/10 & 7/10 \\
Suggested wrong cause selected & -- & 2/10 \\
Lower score than paired control & -- & 3/10 \\
\bottomrule
\end{tabular}
\end{table}

Table~\ref{tab:sycophancy} shows a limited degradation under misleading conjectures. The mean final score decreases from 80.0 to 76.0 and the potential score from 90.0 to 85.0; exact final diagnoses decrease from six to five cases, and exact hypothesis coverage from eight to seven. Thus, misleading suggestions can affect both candidate exploration and final selection, while the correct explanation remains available in most cases.

The explicitly suggested incorrect cause is selected in two of ten challenged cases, and the challenged condition lowers the diagnostic score in three. These results suggest that incorrect user beliefs can influence the reasoning trajectory without systematically anchoring SIA to the proposed explanation. This behavior is consistent with SIA's design goal of retaining user-proposed causes as hypotheses to investigate rather than treating them as evidence by default.

Overall, the paired stress test suggests that SIA generally preserves evidence-driven diagnosis under plausible but misleading user conjectures, while remaining susceptible to some user influence. The test is limited to explicit first-person conjectures and does not cover indirect suggestions, repeated persuasion, or adversarial dialogue~\cite{hong2025measuring}.

\begin{table}[thb!]
\centering
\small
\caption{User feedback on resolving technical problems. Values are average Likert scores on a 1--5 scale.}
\label{tab:user_feedback}
\begin{tabular}{lcccc}
\toprule
Criterion & \texttt{BAS} & \texttt{THK} & \texttt{THK-int} & \texttt{SIA} \\
\midrule
Usefulness & 3.1 & 3.7 & 4.0 & 4.6 \\
Clarity of process & 2.8 & 3.8 & 3.9 & 4.7 \\
Trust in solution & 2.9 & 3.6 & 4.2 & 4.5 \\
Handling ambiguity & 2.8 & 3.5 & 3.7 & 4.5 \\
Perceived time efficiency & 4.8 & 4.3 & 3.2 & 3.0 \\
Overall preference & 3.0 & 3.7 & 4.0 & 4.5 \\
\bottomrule
\end{tabular}
\end{table}

\subsection{User Feedback}

To complement the agent-based evaluation, we conducted a preliminary user study with 10 technically trained participants. Each participant used the \texttt{BAS}, \texttt{THK}, \texttt{THK-int}, and \texttt{SIA} interfaces on self-proposed technical problems and rated each configuration on a 5-point Likert scale. The questionnaire assessed usefulness, process clarity, trust in the final solution, ambiguity handling, perceived time efficiency, and overall preference.

Table~\ref{tab:user_feedback} reports the average ratings. Participants preferred \texttt{SIA} across the main qualitative criteria, with the largest gains observed for clarity of process and ambiguity handling. Users reported that the investigator-style interaction made the diagnostic process easier to follow because the system asked targeted questions, compared alternative explanations, and did not immediately commit to the first plausible cause. Several participants also noted that this behavior increased their trust in the final answer, especially when the initial problem description was incomplete or influenced by their own hypothesis.

The perceived time-efficiency ratings highlight a complementary trade-off. BAS and THK were perceived as faster because they require little or no interaction, whereas THK-int and SIA require additional time to answer clarification questions and complete the diagnostic process. Despite this higher perceived interaction cost, SIA received the highest overall preference. BAS received lower ratings for trust and ambiguity handling, while THK improved several qualitative ratings but still lacked the interactive evidence-gathering behavior of SIA.

\section{Conclusions}
\label{sec:conclusion}

This paper introduced SIA, an evidence-first agent that orchestrates specialized reasoning behaviours for interactive technical diagnosis. Rather than committing early to the user's initial interpretation, SIA maintains competing hypotheses, asks targeted clarification questions, updates their plausibility as new evidence becomes available, and performs evidence-grounded final selection.
Experiments on solved troubleshooting threads show that SIA achieves higher diagnostic quality and candidate-pool potential than the compared baselines. Ablations highlight complementary contributions from structured investigation, hypothesis expansion, follow-up questioning, plausibility updates, counterfactual critique, and final adjudication. SIA also preserves its advantage over THK-int across model families and shows robustness to plausible but incorrect user conjectures. A preliminary user study further reports higher perceived usefulness, clarity, trust, and ambiguity handling despite the additional interaction cost.
Overall, the results support an investigation-oriented approach in which reliable diagnosis depends not only on reasoning over the initial problem, but also on gathering discriminative evidence and preserving competing explanations. Future work will investigate improved final selection, calibrated uncertainty estimation, adaptive stopping strategies, and evaluation on additional diagnostic domains.


\bibliographystyle{plain}
\bibliography{references}

\end{document}